%% file: root.tex
\documentclass[letterpaper, 10 pt, conference]{ieeeconf}  

\IEEEoverridecommandlockouts                              

\usepackage{hyphenat}
\usepackage{graphics} 
\usepackage{graphicx}
\usepackage{float}
\usepackage{booktabs}
\usepackage{multirow}
\usepackage{threeparttable}
\usepackage{algorithm}
\usepackage{algpseudocode}
\usepackage{epsfig} 
\usepackage{times} 
\usepackage{amsmath}
\usepackage{mathrsfs}
\usepackage{caption}
\usepackage{cite}
\usepackage{xcolor}
\usepackage{balance}
\title{\LARGE \bf
MCFR: A Mask-Guided Coarse-to-Fine Regression Framework for Robust Multi-Variant  Board-to-Board Connector Assembly 
}

\author{Guanghui Shen, Song Wang and Dan Wu
\thanks{This work was supported by the National Natural Science Foundation of China under Grant 52375019.}%
\thanks{G. Shen, S. Wang, and D. Wu are with the Department of Mechanical Engineering, Tsinghua University, Beijing, China. Corresponding author: Dan Wu ({\tt\small wud@tsinghua.edu.cn}).}%
}

\begin{document}
\bstctlcite{IEEEexample:BSTcontrol}

\maketitle
\pagestyle{empty}  
\thispagestyle{empty} 
\pagestyle{empty}  
\thispagestyle{empty} 
\thispagestyle{empty}
\pagestyle{empty}

\input{article/abstract}
\input{article/introduction}
\input{article/realted_work}
\input{article/method}
\input{article/experiments}
\input{article/conclusion}

\bibliographystyle{IEEEtranBST/IEEEtran}
\balance
\bibliography{IEEEtranBST/reference}

\end{document}

%% file: article/abstract.tex
\begin{abstract}
Automated insertion of board-to-board (BTB) connectors in 3C manufacturing requires both high visual accuracy and strong deployment robustness. This problem remains challenging because multi-variant connectors exhibit significant morphological and appearance variations, making stable cross-variant generalization difficult, while the mismatch between training and deployment under fixed-view inspection settings induces background spurious correlation and degrades real-world performance. To address these issues, this paper proposes MCFR, a Mask-Guided Coarse-to-Fine Regression framework for multi-variant BTB connector assembly. By introducing an object-aware mask prior and explicit photometric refinement, the proposed method suppresses background interference and improves alignment accuracy and robustness in practical deployment. Experiments on a self-constructed multi-variant dataset, a BTB batch insertion testbed, and a real smartphone assembly task show that MCFR consistently outperforms representative baselines and achieves an average real-world insertion success rate of 99.25\%. These results demonstrate the effectiveness and practical potential of MCFR for automated assembly of multi-variant BTB connectors.

\end{abstract}

%% file: article/introduction.tex
\section{INTRODUCTION}
Within the 3C manufacturing sector (computers, communications, and consumer electronics), automated insertion of board-to-board (BTB) connectors is a critical process for realizing flexible assembly of intelligent terminals\cite{THUWS-RAL, qin2026assemblyreview, yang2025tactilefinger}. BTB connectors are characterized by substantial category diversity, miniature dimensions, pronounced appearance variations, and weak discriminative localization features\cite{RAL-ASSEMBLY, RAL-TACTILE}. In addition, their assembly process imposes extremely stringent tolerance requirements, typically within ±0.2 mm and ±1.5 ° \cite{BUAA-TIE}. Consequently, the accuracy and robustness of relative pose detection directly determine the feasibility of automated insertion for multi-variant BTB connectors, thereby constituting the principal bottleneck that hinders the stable deployment of such automation systems~\cite{li2019survey,jiang2020measurement,xu2023high}.

\begin{figure}[tbp]
\centering
\includegraphics[scale=0.61,trim=2 6 0 0,clip]{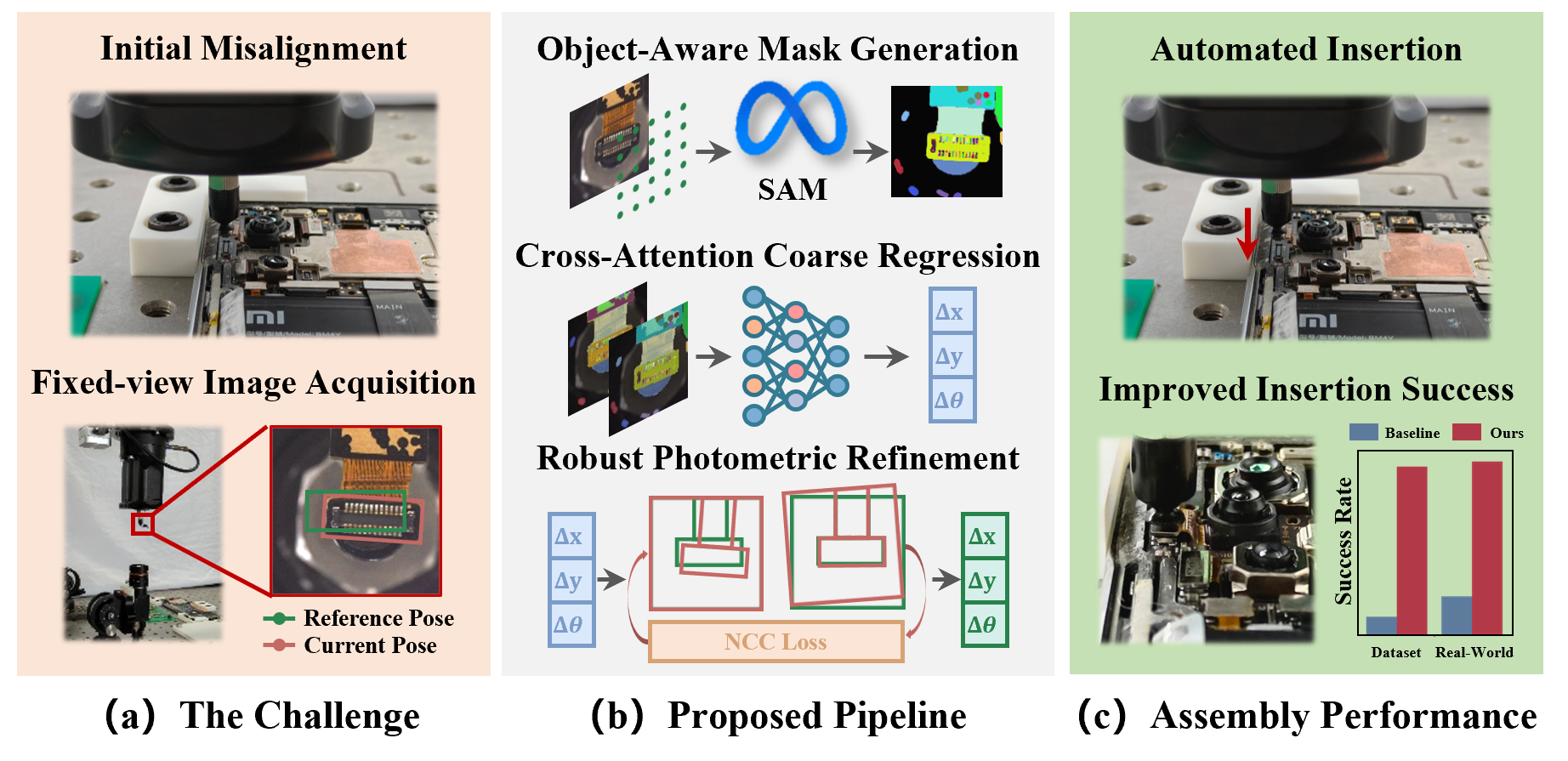}
\caption{Overview of the proposed method. (a) Initial grasping introduces pose misalignment, and fixed-viewpoint images are acquired. (b) The images are fed into the proposed mask-guided coarse-to-fine regression framework for relative pose estimation. (c) The estimated pose is used for insertion correction, improving insertion success in real-world open scenarios.}
\vspace{-0.25cm}
\label{fig:pic1}
\end{figure}

To address this bottleneck, existing studies typically adopt YOLO-based object detection methods \cite{jing2023fpc}, \cite{mou2022pose} or directly employ end-to-end deep networks for relative pose regression \cite{yu2019siamese}, \cite{zhang2023fusing}, \cite{li2024v}. Although these methods have demonstrated a certain degree of effectiveness in controlled scenarios, they still face two critical challenges in real flexible production-line deployment. First, purely data-driven implicit regression models are often highly sensitive to the geometric morphology and appearance distribution of connectors, making it difficult to achieve stable generalization across multiple BTB connector variants with substantial morphological differences. Second, in practical visual inspection settings with a fixed viewpoint, training samples are commonly generated by moving the connector through random displacements using a robotic arm, causing the target and the background to exhibit coupled rigid-body transformations in the image. However, in actual assembly, the initial pose mismatch is mainly induced by grasping errors\cite{li2019survey}, whereas the background usually remains completely static. In this paper, we refer to the phenomenon arising from this mismatch between training and deployment distributions—where the model tends to rely on background displacement cues rather than object-intrinsic features—as background spurious correlation. This issue substantially degrades the generalization performance of end-to-end methods in real-world open scenarios, and can be viewed as a task-specific manifestation of shortcut learning under distribution shift~\cite{geirhos2020shortcut}.

To mitigate this issue, this paper proposes MCFR, a mask-guided coarse-to-fine robust pose regression framework for multi-variant connectors in real-world open scenarios. Rather than relying on direct end-to-end black-box regression, MCFR introduces the visual foundation model Segment Anything Model (SAM) \cite{SAM} to generate an object-aware probability mask. As a visual prior, this mask can explicitly indicate the spatial distribution of the BTB connector and its effective localization features, thereby suppressing background interference at the source. The original image and the mask prior are then jointly fed into a Vision Transformer (ViT)-based pose regression network \cite{ViT} to produce an initial coarse pose estimate. Finally, to ensure robustness under open scenarios and to resist performance degradation caused by real environmental factors, we design an explicit photometric refinement module that performs pixel-level precise alignment of the coarsely regressed pose through robust optimization.

We conduct systematic evaluations of the proposed method on a self-constructed multi-variant BTB connector dataset, a batch insertion test platform, and a real smartphone assembly scenario. Experimental results show that MCFR outperforms existing baseline methods in both relative pose estimation error and real insertion success rate. 

The main contributions of this paper are summarized as follows.

\begin{itemize}
\item We propose MCFR, a robust pose detection framework that overcomes the morphological generalization challenges of multi-variant BTB connectors, providing a reliable visual solution for high-precision physical assembly.
\item We reveal the issue of \textit{background spurious correlation} in fixed-viewpoint relative pose detection and design a mask-guided architecture with explicit photometric refinement to mitigate background reliance and enhance robustness in open scenarios.
\item Extensive experiments on datasets and real-world assembly platforms demonstrate that MCFR achieves superior error metrics and a near-perfect (99.25\%)  insertion success rate, significantly outperforming existing baselines.
\end{itemize}

%% file: article/realted_work.tex
\section{Related Work}
In the field of vision-based flexible robotic assembly, relative pose detection has consistently been a critical issue for achieving high-precision insertion and stable assembly\cite{jiang2020measurement, xu2023high}. To address this problem, existing research can be broadly categorized into two streams: analytical pose estimation methods based on feature engineering, and end-to-end pose regression methods.

\subsection{Analytical Pose Estimation Based on Feature Engineering}
Such methods typically begin by extracting local localization features from images, such as edges, keypoints, or specific color patches. These features are then combined with geometric or statistical analysis algorithms (e.g., Principal Component Analysis (PCA)\cite{jolliffe2016pca}, Linear Discriminant Analysis (LDA)\cite{fukunaga2013spr}, or specific geometric solvers) to calculate the relative pose. Early feature extraction primarily relied on traditional image processing operators like Canny edge detection \cite{bao2005canny} and the Hough transform\cite{duda1972hough}. However, because traditional operators heavily depend on manually set empirical thresholds, their generalization capability and robustness to interference are relatively limited when faced with complex backgrounds or multiple part categories. To overcome this bottleneck, recent studies have increasingly introduced deep learning object detection models, such as YOLO \cite{redmon2016you}, leveraging their powerful high-dimensional feature representation capabilities to extract more robust localization features. This has significantly improved the stability of front-end feature acquisition; nevertheless, their representations are generally biased towards object-level detection, resulting in a relatively limited capacity to describe fine geometric details.

For instance, Wang et al.\cite{wang2025high, wang2025two} proposed 6-D pose estimation methods that extract the upper and lower surface edges of components, combined with Dual-Space Reprojection Closed-Loop Optimization (DSRCEO). While this method achieved high alignment precision in the assembly of commonly shaped thin-plate components, it imposes strict requirements on edge extraction quality and is more suitable for structured scenarios with minimal background interference. Mou et al.\cite{mou2022pose}  presented a framework integrating a YOLO network with manifold learning. By detecting the internal pins of flexible connectors and utilizing their specific geometric layout features to regress the relative pose, they verified the method's effectiveness in insertion tasks for specific connectors.

Overall, feature engineering-based methods are constrained by the design cost and extraction quality of handcrafted features. Particularly when dealing with BTB connectors characterized by a wide variety, significant appearance variations, and a lack of unified, salient geometric features, it is extremely difficult to design a feature engineering pipeline that balances both high precision and strong generalization capabilities.
\subsection{End-to-End Pose Regression and Perception Enhancement}
To reduce reliance on manual feature design, end-to-end methods exploit the feature representation and data fitting capabilities of deep networks to map sensory inputs directly to pose deviations or control commands. These approaches circumvent tedious explicit feature design and demonstrate stronger representational capacity in modeling the complex appearances of objects.

Yu et al.\cite{yu2019siamese} were among the earliest to propose a relative pose regression method based on Siamese Convolutional Neural Networks (Siamese CNN), directly predicting pose deviations from images, thereby validating the feasibility of end-to-end visual regression in low-tolerance connector insertion tasks. Addressing occlusion issues in micro-assembly, Yang et al.\cite{BUAA-TIM} introduced a Digital Twin (DT) mechanism to compensate for physical visual blind spots via virtual multi-viewpoints, enhancing measurement capabilities in occluded scenarios. Furthermore, Yang et al.\cite{BUAA-TIE} modeled Flexible Printed Circuit (FPC) position recognition as a tactile image alignment problem and utilized a Deep Siamese Network (DSN) to achieve relative position regression. Building upon this, Li et al. \cite{li2024v} proposed a visuo-tactile fusion framework, V-HCP, which handles missing modalities through a compensation branch, further enhancing perceptual robustness under complex working conditions.

In summary, end-to-end methods possess distinct advantages in reducing manual design and enhancing representation capabilities for complex scenes. However, most of these methods are constructed for specific object categories and sensor configurations, lacking explicit constraints on spatial geometric relationships. In real production scenarios, when test samples fall out-of-distribution (OOD), or when the system simultaneously faces background variations and transitions between multi-variant objects, these models are still prone to performance degradation. Consequently, they struggle to meet the requirements of flexible production lines for stable deployment and cross-category generalization.

To mitigate the aforementioned shortcomings, this paper focuses on the robust, high-precision pose detection of multi-variant BTB connectors under a fixed viewpoint. Distinct from direct end-to-end pose regression methods, we propose a three-stage framework combining object-aware mask priors, coarse regression, and explicit optimization. This approach aims to simultaneously enhance cross-category generalization, open-scenario robustness, and high-precision alignment capabilities\cite{chang2024insertone}.

%% file: article/method.tex
\section{Method}
\subsection{Problem Statement}
This paper considers the problem of vision-based relative pose detection for automated insertion of multi-variant BTB connectors. In a typical assembly pipeline, the robot sequentially performs three stages: recognition and grasping, detection and alignment, and controlled insertion. Since the grasping stage involves relatively large-range motion, the in-hand pose of the connector is inevitably affected by robot positioning error and visual recognition error. As a result, an additional detection-and-compensation step is required before insertion to ensure accurate alignment and stable assembly.

Different from pose estimation in visual navigation, which estimates the absolute pose of the camera with respect to the scene\cite{kendall2015posenet,mur2017orbslam2}, the relative pose considered in this work is defined as the pose difference of the object between two inspection images acquired after two grasping actions. This formulation is motivated by the tight mating requirement of BTB connectors. The typical mating tolerance is approximately $\pm 0.2$ mm, while the empirical absolute positioning error of the robot is around $\pm 0.1$ mm, which is already comparable to the assembly tolerance. If absolute pose estimation is directly used for insertion, the visual measurement error and the robot absolute positioning error will accumulate, thereby substantially reducing the insertion success rate. By contrast, the robot repeatability is about $\pm 0.03$ mm. Therefore, instead of solving an absolute localization problem, we reformulate the task as relative pose detection under a fixed-view inspection setting and exploit repeatable positioning together with pose compensation to achieve higher practical alignment accuracy.

Specifically, before deployment, the robot is first guided by demonstration to complete one successful insertion, and the connector image captured at the inspection station under the nominal grasp is recorded as the reference image, denoted by $I_r$. During subsequent operation, after each grasp, the robot moves the connector to the same fixed inspection station and captures the current image, denoted by $I_s$. The objective is to estimate the planar relative pose of the current image with respect to the reference image,
\begin{equation}
\Delta \mathbf{p} = [\Delta x, \Delta y, \Delta \theta]^{\top},
\end{equation}
where $\Delta x$ and $\Delta y$ denote in-plane translational offsets and $\Delta \theta$ denotes the rotational offset about the optical axis. Through the calibrated coordinate transformation, this relative pose is converted into an insertion-stage compensation command for the robot end-effector, thereby enabling pose correction prior to mating and stable insertion.

\subsection{MCFR: Mask-Guided Coarse-to-Fine Regression}

Given a reference image $I_r$ and a current image $I_s$, the proposed MCFR estimates their relative pose in three stages: 1) object-aware mask generation, 2) cross-attention coarse regression, and 3) robust photometric refinement. This design decomposes the pose detection process into visual-prior construction, initial regression, and explicit optimization, thereby jointly improving cross-category generalization, robustness in open scenarios, and final alignment accuracy.

\begin{figure*}[htbp]
\centering
\includegraphics[scale=0.9]{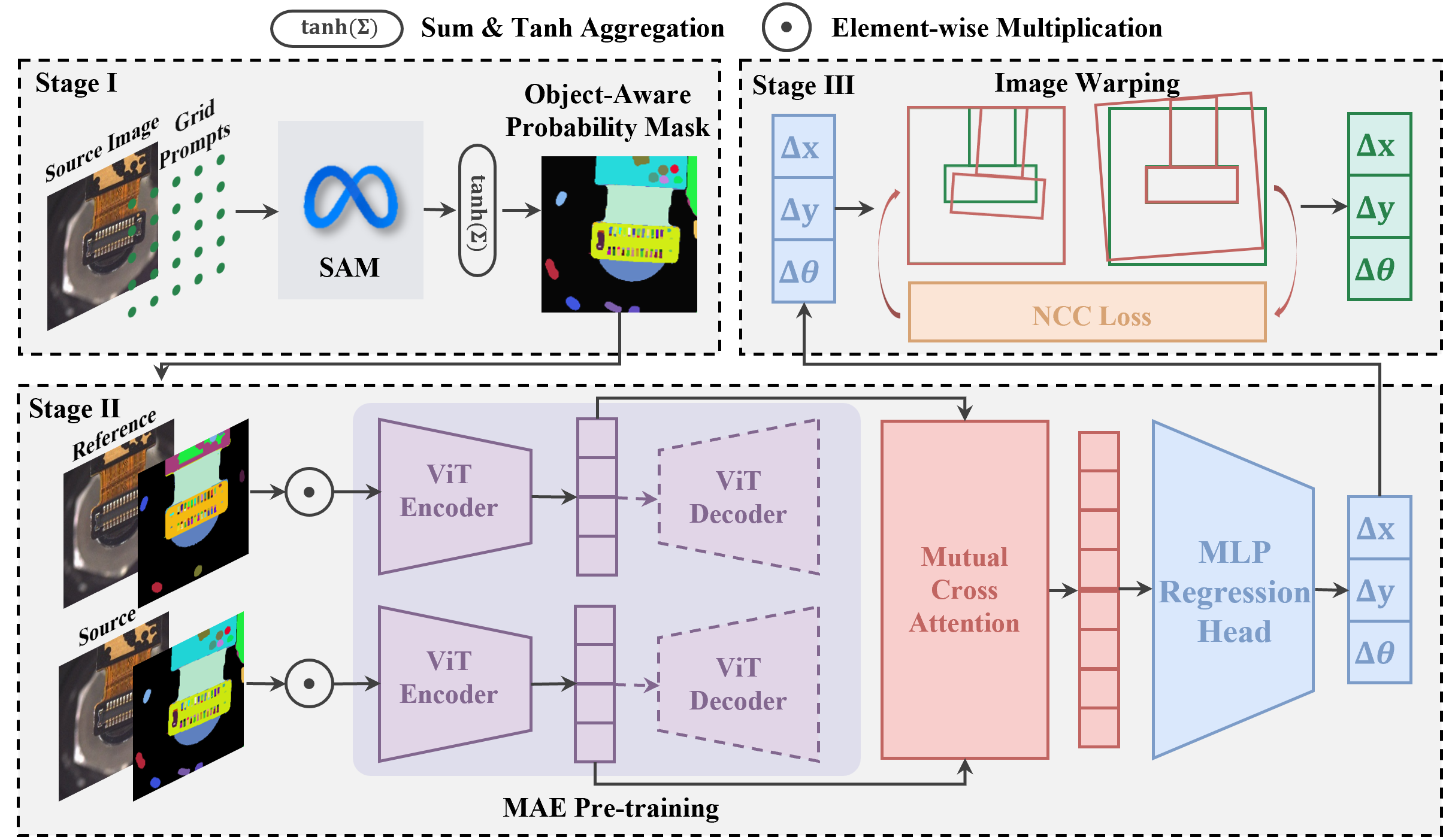}
\caption{Three-stage architecture of MCFR. Stage I constructs an object-aware probability prior from SAM responses to suppress background interference. Stage II performs mask-guided coarse regression to estimate an initial relative pose. Stage III refines the estimate through image-domain photometric alignment, yielding the final relative pose.}
\vspace{-0.25cm}
\label{fig:framework}
\end{figure*}

\subsubsection{Object-Aware Mask Generation (OAMG)}
To mitigate the background spurious correlation discussed in the Introduction and to improve the network's focus on discriminative localization cues, we first construct an object-aware mask as a visual prior. Specifically, we introduce SAM to produce target probability responses without task-specific fine-tuning. For each input image, uniformly distributed grid prompts are placed over the image plane, and the image together with these prompts is fed into SAM. For each prompt point, SAM returns a probability map indicating the likelihood of target presence.

Since some prompts may activate background or irrelevant regions, we further apply a rectangular filtering strategy to discard invalid candidates. Specifically, for each candidate probability map, we extract its dominant contour from the response region and compute a rectangularity score
\begin{equation}
\rho_i = \frac{A_i}{A_i^{\mathrm{rect}}},
\end{equation}
where $A_i$ is the contour area and $A_i^{\mathrm{rect}}$ is the area of its minimum-area bounding rectangle. Only maps with sufficiently high rectangularity and spatial extent consistent with the connector body are retained, since task-relevant localization cues under the fixed-view setting are typically distributed in an approximately rectangular region. The retained maps are then aggregated by pixel-wise summation followed by $\tanh(\cdot)$ normalization:
\begin{equation}
M = \tanh \left( \sum_{i \in \Omega} M_i \right),
\end{equation}
where $M_i$ denotes the probability map of the $i$-th prompt, and $\Omega$ is the valid set after filtering. The resulting mask explicitly highlights the connector body and its effective localization regions, thereby suppressing background interference at the input level.

\subsubsection{Cross-Attention Coarse Regression (CACR)}

After obtaining the input image and its object-aware mask, we apply the mask to the image through pixel-wise multiplication to enhance target responses while suppressing the background:
\begin{equation}
\tilde{I}_r = I_r \odot M_r, \qquad \tilde{I}_s = I_s \odot M_s,
\end{equation}
where $\odot$ denotes element-wise multiplication, and $M_r$ and $M_s$ are the masks corresponding to the reference and current images, respectively.

The two masked images are then encoded by a ViT-based backbone to extract global visual representations. Compared with convolutional architectures, ViT offers better structural scalability and is more suitable for handling the appearance variations introduced by multi-variant connectors. To improve training efficiency and feature quality, we pre-train the ViT encoder using the Masked Autoencoder (MAE) framework\cite{MAE}.

After feature extraction, a mutual cross-attention module is employed to explicitly model the correspondence between the reference and current features, thereby enhancing sensitivity to relative geometric differences. The fused features are then fed into an MLP regression head to obtain an initial estimate of the relative pose:
\begin{equation}
\hat{\Delta \mathbf{p}}_0 = [\hat{\Delta x}_0, \hat{\Delta y}_0, \hat{\Delta \theta}_0]^{\top}.
\end{equation}
This stage leverages the fitting capability of deep networks to provide a stable initialization for subsequent refinement.

\subsubsection{Robust Photometric Refinement (RPR)}
Although the coarse regression network provides an effective initial estimate, its performance may still degrade in real deployment due to illumination variation, background disturbance, and cross-category distribution shift. To further improve robustness in open scenarios and enhance the final detection accuracy, we introduce an explicit photometric refinement module on top of the coarse prediction.

Specifically, the coarse estimate $\hat{\Delta \mathbf{p}}_0$ is used as the initialization for optimization, and the corresponding planar rigid-body transformation is applied to the current image in the image domain.  Let
\begin{equation}
I_w = \mathcal{W}(I_s; \Delta \mathbf{p}),
\end{equation}
where $\mathcal{W}(\cdot)$ denotes the SE(2) image warping parameterized by the relative pose. Intuitively, a more accurate relative pose should yield higher photometric consistency between the warped current image $I_w$ and the reference image $I_r$. Following the normalized cross-correlation (NCC) formulation~\cite{lewis1995fast}, the similarity between $I_r$ and $I_w$ is computed as
\begin{equation}
\frac{\sum\limits_{u \in \mathcal{R}} \left(I_r(u)-\bar{I}_r\right)\left(I_w(u)-\bar{I}_w\right)}
{\sqrt{\sum\limits_{u \in \mathcal{R}} \left(I_r(u)-\bar{I}_r\right)^2}
\sqrt{\sum\limits_{u \in \mathcal{R}} \left(I_w(u)-\bar{I}_w\right)^2}} \, ,
\end{equation}
where $\mathcal{R}$ denotes the valid image region, and $\bar{I}_r$ and $\bar{I}_w$ are the mean intensities of $I_r$ and $I_w$ over $\mathcal{R}$, respectively. The optimization objective is then defined as
\begin{equation}
\mathcal{L}_{\mathrm{NCC}} = 1 - \mathrm{NCC}(I_r, I_w).
\end{equation}
Compared with direct pixel-wise error, NCC is more robust to global illumination shifts and local contrast changes.

After optimization, the refined relative pose $\hat{\Delta \mathbf{p}}$ is obtained and used for pose compensation before insertion. By combining deep regression with explicit optimization, MCFR preserves cross-category adaptability while further improving detection stability and alignment precision in real assembly scenarios.

%% file: article/experiments.tex
\section{Experiments}
\subsection{Overview}
Experiments are conducted from three aspects: dataset-level evaluation, real-world insertion testing on the BTB mating testbed, and a practical smartphone assembly task. On the self-constructed multi-variant dataset, the proposed method is compared with representative baselines, together with ablation studies and occlusion sensitivity visualization. On the BTB mating testbed, batch insertion experiments on multiple connector variants are performed to evaluate real-world insertion success rates. Finally, a real smartphone and camera-module insertion task is used to verify the feasibility of the proposed method in a complete assembly pipeline.

\subsection{Experimental Setup}

\subsubsection{Hardware}
\begin{figure}[!ht]
\centering
\includegraphics[scale=0.72,trim=0 0 0 0,clip]{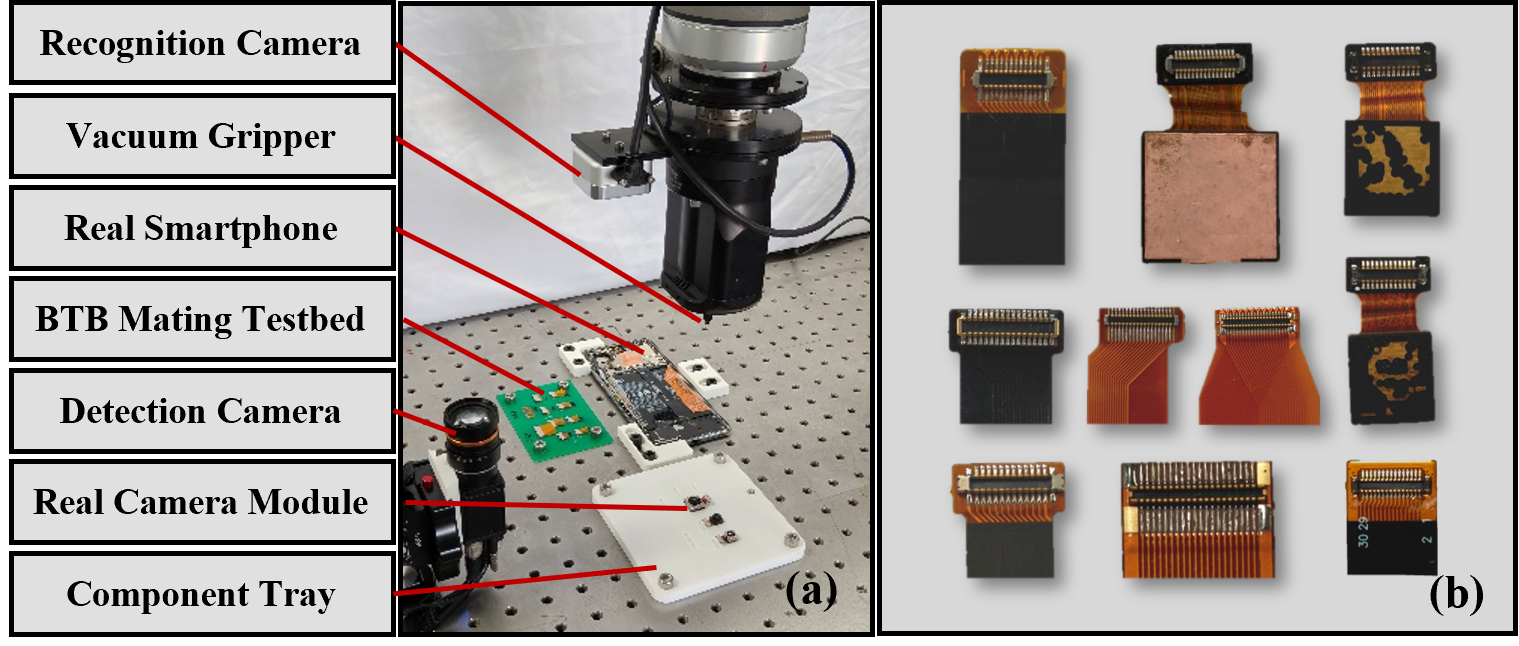}
\caption{Experimental platform and representative test objects. (a) Real assembly setup, including the recognition camera, vacuum gripper, BTB mating testbed, detection camera, real smartphone, and component tray. (b) Representative BTB connectors and smartphone camera-module components used in the experiments.}
\vspace{-0.25cm}
\label{fig:hardware}
\end{figure}

The experimental platform is shown in Fig. \ref{fig:hardware}. A UR5e manipulator is used as the robotic platform, with a repeatability of $\pm 0.03$ mm. An EPick vacuum gripper is adopted to grasp miniature connector components. An Intel RealSense D405 is used for grasp-stage visual localization, while a Hikvision MV-CS060-10GC monocular camera with a resolution of $3072 \times 2048$ is used for high-precision relative pose detection. A BTB mating testbed that integrates 10 connector sockets is constructed for batch insertion experiments. In addition, a Redmi K40 smartphone and camera modules are used to validate the proposed method in a realistic assembly scenario. A component tray is used for the initial placement and grasping of parts. Fig. \ref{fig:hardware}(b) shows representative BTB connectors and camera-module components used in the experiments.

\subsubsection{Dataset Collection}
The dataset is collected by grasping each part with the robot and randomly perturbing its pose within the range of $[\pm 4.5\ \mathrm{mm}, \pm 3\ \mathrm{mm}, \pm 10^{\circ}]$. The dataset includes 10 BTB connector variants and 3 smartphone camera modules, with 300 image pairs collected for each variant. To improve robustness under varying illumination, we additionally collect 300 image pairs under weak lighting and 300 image pairs under strong lighting for a yellow-plated BTB connector featuring 24 pins with a pin pitch of 0.35 mm. The robot relative motion is used as the ground-truth relative pose label. In total, the dataset contains 4500 samples, which are split into the training, validation, and test sets with a ratio of 8:1:1.

To evaluate the impact of background spurious correlation, the original test split is further post-processed to construct a simulated deployment test set. Specifically, the connector region is first segmented from each test image and then composited onto a separately captured empty background image of the fixed inspection station. As a result, only the connector pose varies while the background remains fixed, which better matches the visual condition in real deployment after grasping.

\subsubsection{Model Configuration and Training}
All experiments are conducted on a workstation equipped with an Intel Core i7-13700KF CPU and an NVIDIA GeForce RTX 4080 Super GPU. The encoder adopts a ViT-Base architecture. Before supervised training, the encoder is pre-trained in a self-supervised manner using the MAE framework on the collected images to improve scene representation. To reduce GPU memory usage during training, SAM is used only for offline preprocessing to generate object-aware masks, which are saved as RGB images. The CACR network is then trained with shared encoder weights for the two branches and supervised by an MSE loss. The supervised training of the CACR network runs for 100 epochs and takes approximately 22 minutes.

\subsection{Algorithm Comparison and Ablation Study}\label{subsec:alg_comp}
We evaluate all methods using mean absolute translation error (Trans. MAE), mean absolute rotation error (Rot. MAE), and their corresponding maximum errors. Two representative baselines are implemented for comparison. \textbf{YOLO+LDA:} following Mou et al.~\cite{mou2022pose}, this analytical pipeline combines YOLO-based feature extraction with LDA-based geometric analysis. Since many connectors share yellow pin regions, it first segments the yellow pin array using YOLO, then computes the localization center from the centroids of the segmented pin masks, and finally estimates the rotation through LDA-based manifold analysis. \textbf{ResNet50+LSTM:} implemented following Yang et al.~\cite{BUAA-TIE}, this Siamese end-to-end regression model uses a pre-trained ResNet50 as the image encoder, and regresses the concatenated feature vectors with an LSTM head.

\begin{table*}[t]
\centering
\caption{Quantitative comparison of different methods on the validation and test sets.}
\label{tab:quantitative_comparison}
\setlength{\tabcolsep}{4pt}
\renewcommand{\arraystretch}{1.15}
\scriptsize
\begin{tabular}{l|cccc|cccc}
\toprule
\multirow{2}{*}{Model}
& \multicolumn{4}{c|}{Validation Set}
& \multicolumn{4}{c}{Test Set} \\
\cmidrule(lr){2-5} \cmidrule(lr){6-9}
& Trans. MAE (mm)
& Rot. MAE ($^\circ$)
& Trans. Max. (mm)
& Rot. Max. ($^\circ$)
& Trans. MAE (mm)
& Rot. MAE ($^\circ$)
& Trans. Max. (mm)
& Rot. Max. ($^\circ$) \\
\midrule
YOLO+LDA      & 0.095 & 0.117
 & 0.283 & 0.331 & 0.094 & 0.142 & 0.280 & \textbf{0.428} \\
ResNet50+LSTM   & 0.103 & 0.316 & 0.339 & 1.027 & 1.062 & 1.335 & 3.690 & 4.712 \\
MCFR w/o OAMG & 0.021 & 0.047 & 0.076 & \textbf{0.213} & 0.793 & 0.192 & 3.049 & 1.138 \\
MCFR w/o RPR  & 0.021 & 0.025 & 0.159 & 0.245 & 0.052 & 0.257 & 0.533 & 1.295 \\
MCFR (Ours)   & \textbf{0.013} & \textbf{0.015} & \textbf{0.064} & 0.498 & \textbf{0.027} & \textbf{0.126} & \textbf{0.196} & 0.833 \\
\bottomrule
\end{tabular}
\end{table*}

\begin{figure}[!ht]
\centering
\includegraphics[scale=0.31,trim=0 8 0 4,clip]{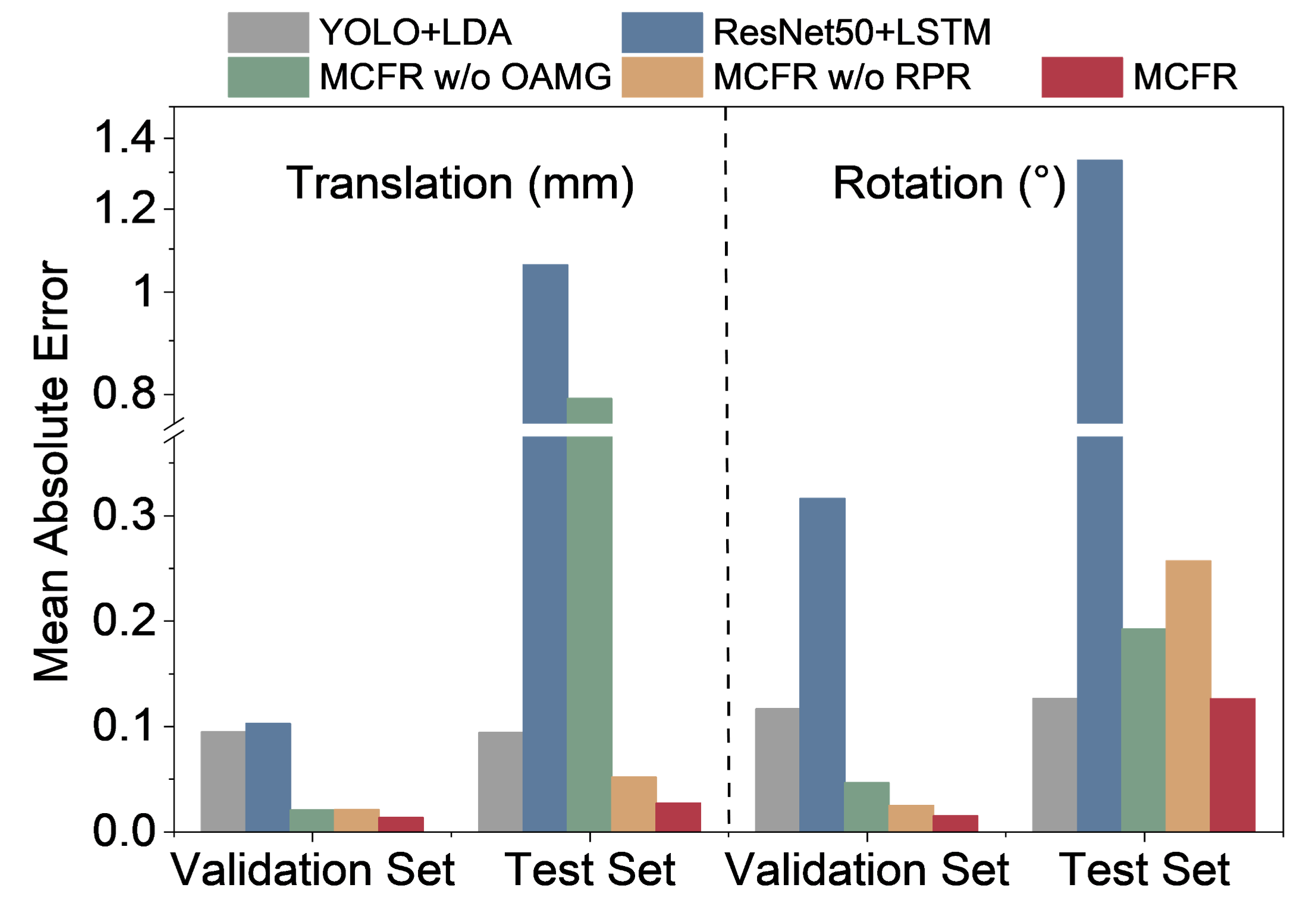}
\caption{Comparison of mean errors on the validation and test sets. The figure reports the performance changes under the fixed-background test setting and the stability advantage of the proposed method.}
\vspace{-0.25cm}
\label{fig:error}
\end{figure}
We further conduct two ablation studies. \textbf{MCFR w/o OAMG:} MCFR without object-aware mask generation. \textbf{MCFR w/o RPR:} MCFR without robust photometric refinement. As shown in Table~\ref{tab:quantitative_comparison} and Fig.~\ref{fig:error}, the proposed MCFR achieves the lowest mean translation and rotation errors on both the validation and test sets. On the validation set, MCFR achieves a Trans. MAE of 0.013 mm and a Rot. MAE of 0.015$^\circ$, which are lower than those of all baselines. On the test set, it still maintains the best mean performance, with a Trans. MAE of 0.027 mm and a Rot. MAE of 0.126$^\circ$, while ResNet50+LSTM degrades to 1.062 mm and 1.335$^\circ$, respectively. In addition, MCFR also achieves the smallest maximum translation errors on both sets, i.e., 0.064 mm on the validation set and 0.196 mm on the test set.

The ablation results further validate the necessity of both modules. In particular, when the object-aware mask is removed, the test Trans. MAE increases sharply from 0.027 mm to 0.793 mm, indicating that the model becomes highly sensitive to the fixed-background deployment setting. By contrast, the full model maintains stable accuracy, showing that the mask prior effectively mitigates background spurious correlation by encouraging the network to rely on connector-intrinsic cues rather than background displacement. Moreover, removing the robust photometric refinement module leads to a noticeable degradation from 0.027 mm to 0.052 mm in test Trans. MAE and from 0.126$^\circ$ to 0.257$^\circ$ in test Rot. MAE, confirming the importance of explicit optimization for accurate and robust deployment.

\begin{figure}[!ht]
\centering
\includegraphics[scale=1.028,trim=0 0.5 0 5,clip]{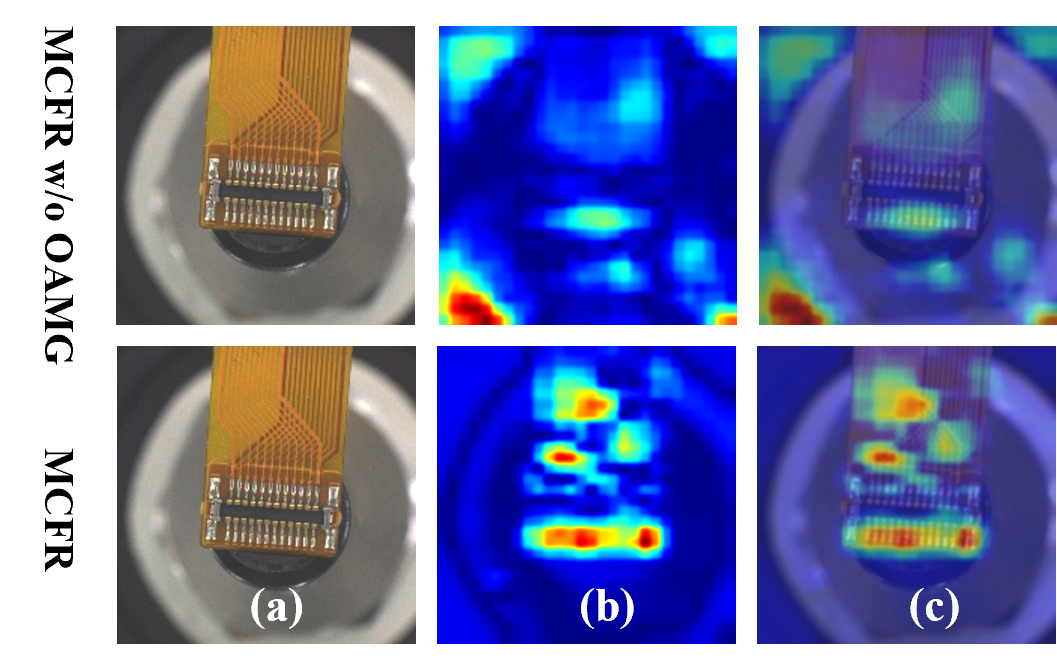}
\caption{Occlusion sensitivity visualization. (a) Input image. (b) Occlusion sensitivity map. (c) Overlay result. The top row corresponds to the model without the object-aware mask, while the bottom row corresponds to the model with the mask prior, illustrating the difference in attention regions.}
\vspace{-0.25cm}
\label{fig:attention}
\end{figure}

In addition to quantitative evaluation, we further visualize the model attention through occlusion sensitivity maps~\cite{zeiler2014visualizing}. Specifically, local image patches are occluded and the resulting prediction degradation is measured to estimate the contribution of each pixel region. The visualization shows that, with the object-aware mask, the network mainly focuses on the BTB connector body, with the highest contribution concentrated around the pin region. By contrast, the model without the mask tends to rely more on image boundaries and the circular outer contour of the gripper in the background. This observation is consistent with the quantitative ablation results and further verifies that the proposed mask prior suppresses background interference and improves deployment robustness.

\subsection{Real-World Insertion Experiments}

\begin{figure*}[t]
\centering
\includegraphics[scale=0.95,trim=0 0 0 0,clip]{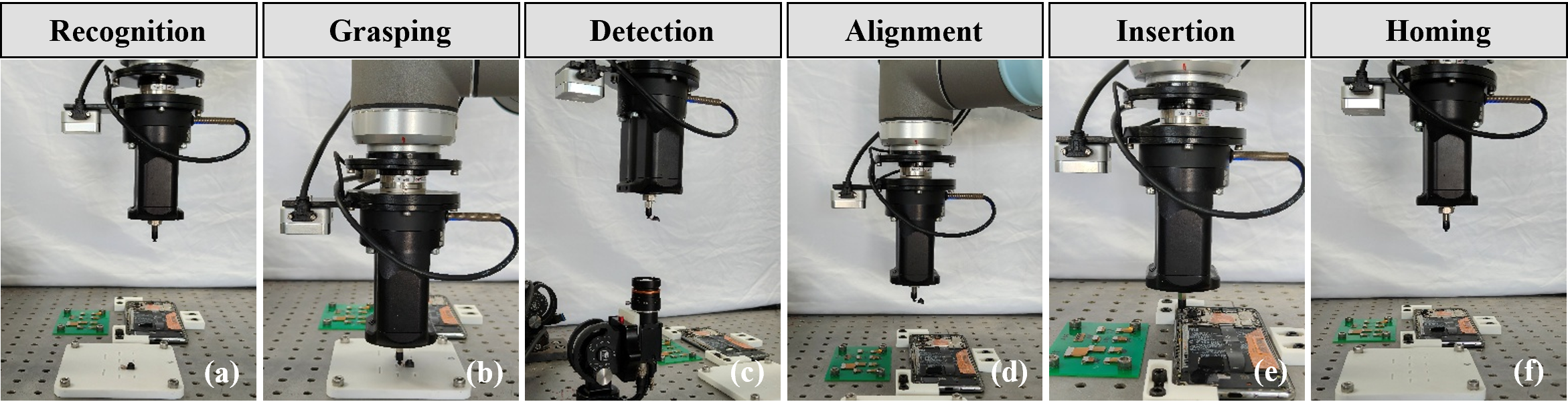}
\captionsetup{justification=raggedright, singlelinecheck=false}
\caption{Complete execution pipeline in the real smartphone assembly task, including recognition, grasping, detection, alignment, insertion, and homing, demonstrating the closed-loop deployment of the proposed method in a practical task.}
\label{fig:real-world}
\vspace{-0.25cm}
\end{figure*}
To evaluate the practical deployment performance of the proposed method, we further conduct real-world insertion experiments, including batch BTB insertion tests and smartphone assembly tests. In this subsection, we focus on the most representative end-to-end baseline and the practically deployable variants of the proposed framework, namely the \textbf{ResNet50+LSTM} baseline, \textbf{MCFR w/o RPR}, and the complete \textbf{MCFR}. A trial is counted as successful if the connector is fully assembled before the contact force along the insertion direction reaches 25~N. This force threshold is set according to the average peak insertion force observed when the connector is manually aligned and inserted successfully.

\begin{figure}[tbp]
\centering
\includegraphics[scale=0.9,trim=0 0 0 0,clip]{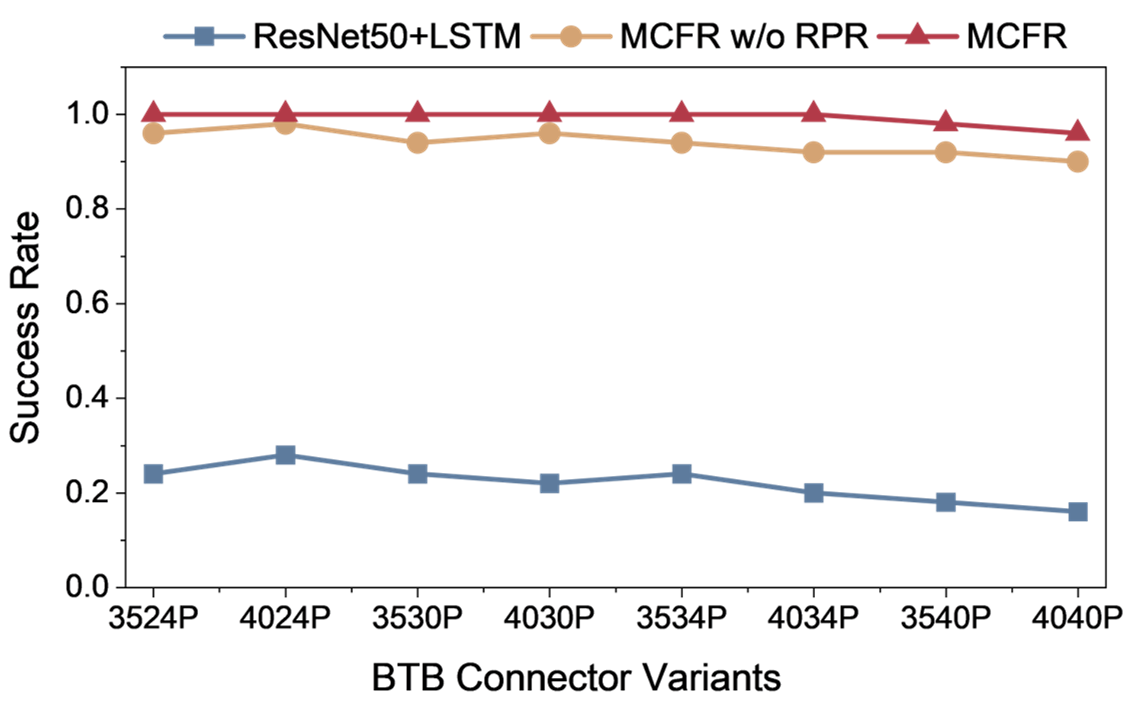}
\caption{Real-world insertion success rates of three methods on 8 BTB connector variants. The horizontal-axis labels encode the pin pitch and pin count of each connector variant, reflecting the difference in assembly difficulty across specifications.}
\vspace{-0.25cm}
\label{fig:real-world-insertion}
\end{figure}

First, batch insertion experiments are conducted on the BTB mating testbed using 8 BTB connector variants. For each category, 50 full trials are performed, and the real-world insertion success rate is recorded. Fig.~\ref{fig:real-world-insertion} compares the success rates of the three methods across the 8 connector variants. The horizontal-axis labels indicate the connector specifications: the first two digits denote the pin pitch (e.g., 35 represents 0.35~mm), while the suffix ``P'' denotes the number of pins (e.g., 24P indicates 24 pins). As the pin count increases, the connector typically becomes more slender and is more prone to jamming under slight alignment errors, thereby imposing stricter requirements on pose estimation and compensation accuracy.

The results show that the average insertion success rates of the three methods over the 8 connector variants are 22\%, 94\%, and 99.25\%, respectively. In particular, the ResNet50+LSTM baseline performs substantially worse across all connector variants, and its success rate drops more noticeably on connectors with larger pin counts. By contrast, MCFR w/o RPR already maintains a high success rate for most connector variants, while the complete MCFR consistently achieves the best performance across all tested connectors. These results indicate that the proposed method not only improves offline pose estimation accuracy but also translates this advantage into significantly higher success rates in practical insertion tasks.

The main reason for this performance gap is the mismatch between real deployment conditions and the training data distribution. In the actual inspection station, the background remains fixed, whereas the training data are mainly collected by moving the grasped part with random robot perturbations, which can cause end-to-end regression models to rely on background displacement cues. When the deployment background no longer follows the training distribution, implicit regression models such as ResNet50+LSTM are more prone to performance degradation. In contrast, MCFR suppresses background spurious correlation through the object-aware mask prior and further improves alignment precision through explicit photometric refinement, resulting in stronger stability and robustness in real deployment.

Finally, we further validate the transferability of the proposed method in a real smartphone camera-module insertion task. The complete pipeline, including recognition, grasping, detection, alignment, insertion, and homing, is successfully executed, as shown in Fig.~\ref{fig:real-world}. Despite the tighter structural constraints and higher assembly difficulty in the real smartphone scenario, the proposed method remains effective for relative pose detection and insertion correction, demonstrating its feasibility in practical deployment for intelligent terminal assembly.

%% file: article/conclusion.tex
\section{CONCLUSIONS}
This paper addresses high-precision relative pose detection and automated insertion for multi-variant BTB connectors under a fixed-view inspection setting, and proposes MCFR, a mask-guided coarse-to-fine regression framework. By introducing an object-aware mask prior, mutual cross-attention coarse regression, and explicit photometric refinement, the proposed method suppresses background spurious correlation while improving alignment accuracy and deployment robustness.

Experiments on the self-constructed multi-variant dataset show that MCFR outperforms representative analytical and end-to-end baselines in both translation and rotation accuracy. Ablation studies and occlusion sensitivity visualization further validate the effectiveness of the proposed design from both quantitative and feature-attribution perspectives. In batch real-world experiments on the BTB mating testbed, MCFR achieves an average insertion success rate of 99.25\% over 8 connector variants, while a real smartphone assembly task further demonstrates the feasibility of the complete pipeline in practical deployment. Although this study is motivated by relative pose detection for BTB connector insertion, the proposed method is not restricted to this specific application. As a general mask-guided coarse-to-fine regression framework, MCFR can be readily extended to other object categories without architectural modification, providing a practical and transferable visual solution for high-precision robotic assembly.

Although MCFR improves cross-variant generalization, zero-shot initialization on entirely unseen connector variants is not explicitly addressed in the current study and remains an important direction for future work.